\documentclass[10pt,twocolumn,letterpaper]{article}

\usepackage[pagenumbers]{cvpr}  
\usepackage[accsupp]{axessibility} 
\usepackage{graphicx}
\usepackage{amsmath}
\usepackage{amssymb}
\usepackage{booktabs}
\usepackage{bm}
\usepackage{comment}  
\usepackage{pifont}
\usepackage{enumitem}
\usepackage[pagebackref,breaklinks,colorlinks]{hyperref}
\usepackage{multirow}

\usepackage[capitalize]{cleveref}
\crefname{section}{Sec.}{Secs.}
\Crefname{section}{Section}{Sections}
\Crefname{table}{Table}{Tables}
\crefname{table}{Tab.}{Tabs.}

\def\cvprPaperID{*****} 
\def\confName{CVPR}
\def\confYear{2025}

\begin{document}

\title{NeISF++: Neural Incident Stokes Field for Polarized Inverse Rendering of Conductors and Dielectrics}

\author{Chenhao Li$^{1, 3}$,\quad Taishi Ono$^2$,\quad Takeshi Uemori$^1$,\quad Sho Nitta$^1$, \\  Hajime Mihara$^1$,  \quad Alexander Gatto$^2$,\quad Hajime Nagahara$^3$,\quad Yusuke Moriuchi$^1$\\
\\
{Sony Semiconductor Solutions Corporation$^1$,\quad Sony Europe B.V.$^2$,\quad  Osaka University$^3$}
}

\twocolumn[{
        \maketitle
        \vspace{-2.5em}
	\begin{center}
            \captionsetup{type=figure}
		\includegraphics[width=\textwidth]{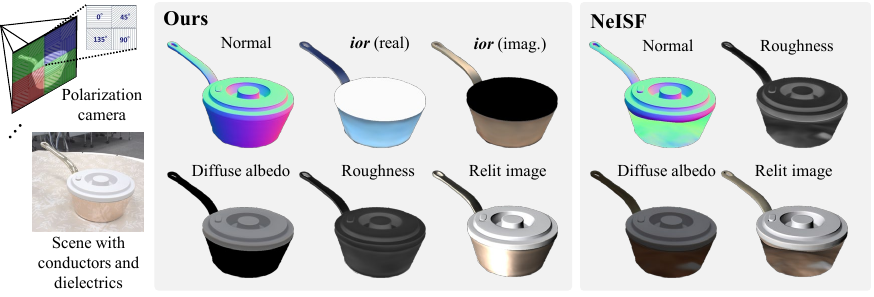}
		\captionof{figure}{Comparison of the polarized inverse rendering methods. Since NeISF \cite{li2024neisf} does not model conductors correctly, its geometry and material reconstruction is poor. Our method estimates the complex refractive index of conductors and re-renders the image via a physically-based pBRDF. Therefore, the reconstructed material and geometry are accurate, and the relighting result is glossy. We normalize the real and imaginary parts of the complex refractive index $\bm{ior}$, and visualize them separately.}	
	\label{fig:teaser}
        \end{center}
}]

\begin{abstract}
Recent inverse rendering methods have greatly improved shape, material, and illumination reconstruction by utilizing polarization cues.
However, existing methods only support dielectrics, ignoring conductors that are found everywhere in life. Since conductors and dielectrics have different reflection properties, using previous conductor methods will lead to obvious errors. In addition, conductors are glossy, which may cause strong specular reflection and is hard to reconstruct.
To solve the above issues, we propose NeISF++, an inverse rendering pipeline that supports conductors and dielectrics. The key ingredient for our proposal is a general pBRDF that describes both conductors and dielectrics. As for the strong specular reflection problem, we propose a novel geometry initialization method using DoLP images. This physical cue is invariant to intensities and thus robust to strong specular reflections.
Experimental results on our synthetic and real datasets show that our method surpasses the existing polarized inverse rendering methods for geometry and material decomposition as well as downstream tasks like relighting. 
\end{abstract}

\section{Introduction}
Inverse rendering is a fundamental task in computer vision and computer graphics, which aims to decompose the target scene into 3D properties like geometry, material, and lighting. It is crucial for applications like virtual reality, material science, and game design. The recent progress of inverse rendering has been dominated by neural representations \cite{mildenhall2020nerf}, which utilize multilayer perceptrons (MLPs)  to efficiently represent geometry \cite{yariv2023bakedsdf, wang2021neus}, material \cite{boss2021nerd}, and lighting \cite{yao2022neilf}, and greatly improve the reconstruction accuracy. However, the inherent challenge in inverse rendering, the ambiguity problem, still exists. Recovering 3D properties from 2D images is essentially an ill-posed problem, as different combinations of geometry, material, and lighting may result in the same appearance.

An active research field for reducing ambiguity is applying neural representations beyond conventional cameras. Advanced sensors such as event \cite{rudnev2023eventnerf, qi2023e2nerf},  infrared \cite{ye2024thermal}, hyper-spectral \cite{poggi2022xnerf},  fisheye \cite{liao2024fisheye}, and time-of-flight cameras \cite{attal2021torf} have been extensively explored. One sophisticated sensor worth mentioning is the polarization camera, which can capture the oscillation direction of light in addition to intensity and color. When the light interacts with the object's surface, the polarization changes according to the geometry and material. In other words, the captured polarization image contains rich information about geometry and material, thus disambiguating the inverse rendering. To our knowledge, PANDORA \cite{dave2022pandora} is the first work that successfully combines polarization cues and neural representations for an inverse rendering problem. Since then, many follow-up polarized inverse rendering works have been proposed to improve it by supervising with tangent space consistency \cite{cao2023multi, han2024nersp}, specifically designed polarimetric loss \cite{chen2024pisr}, or considering indirect illumination \cite{li2024neisf}. However, the same problem with the above polarized inverse rendering works is that they only support dielectric materials. Although dielectrics like rubber, wood, and plastic are common materials, conductors like steel, gold, and aluminum are also unignorable. Applying dielectrics-based inverse rendering methods to conductors causes significant errors. The error mainly comes from two aspects. The first one is dielectrics and conductors have different polarimetric properties, they thus require different material models. The second one is conductors are usually glossy and have strong specular reflections, which increases the ambiguity and requires special treatment.

 To solve these issues, we propose NeISF++, a polarized inverse rendering method that supports both conductors and dielectrics. Our framework mainly follows NeISF \cite{li2024neisf}. Given multi-view polarized images, we represent the geometry as a signed distance field (SDF), the material as a  Bidirectional Reflectance Distribution Function (BRDF) field, and the multi-bounced polarized light as an incident Stokes field. Then, a physically-based polarimetric renderer calculates the final outgoing polarized light, and the model is self-supervised. This work focuses on solving the briefly mentioned two error sources: the material model and the strong specular reflection. For the first error source (material model), we propose a general polarimetric BRDF (pBRDF) that describes both conductors and dielectrics. Existing polarized inverse rendering works \cite{li2024neisf, dave2022pandora} use Baek pBRDF \cite{baek2018simultaneous} as the material model, which is specially designed for dielectrics. It describes the captured polarization signal as the combination of diffuse and specular polarization. The diffuse polarization comes from multi-bounced subsurface scattering, and the specular polarization comes from single-bounced mirror reflection. Extending Baek pBRDF to support conductors faces many challenges. For example, visible light can not penetrate the surface of conductors \cite{collett2005field}. Therefore, the diffuse polarization does not exist for conductors. To solve this discrepancy, we propose using a binary indicator to control the existence of the diffuse polarization term. In addition, even for the single-bounced mirror reflection, the properties of conductors and dielectrics are still different. Because the refractive index of dielectrics is a real number, while the refractive index for conductors is a complex number, we implement a general Fresnel reflection term that supports both real and complex numbers to address this problem. For the second error source (strong specular reflection), we propose using the degree of linear polarization (DoLP) images to initialize the SDF. Training the SDF from the initialization of volume rendering works like VolSDF \cite{yariv2021volume} is a common technique used in the existing inverse rendering works \cite{zhang2023neilf++, li2024neisf}. However, poor geometry initialization can damage the final reconstruction results, and the initialization quality is usually low when strong specular reflection exists. An advantage of using polarized images is that we can utilize physical properties such as DoLP, which is independent of the light intensity and strongly related to the geometry. With these advantages, we argue that DoLP is a better image domain for geometry initialization than intensity images. By solving the two error sources, NeISF++ significantly improves geometry and material reconstruction when both conductors and dielectrics exist. Additionally, because of the correct modeling of conductors, the relighting results are much more realistic than the previous work \cite{li2024neisf} (Fig. \ref{fig:teaser}). 
 To summarize, our contributions can be seen as proposing:
\begin{itemize}[label={$\bullet$}, nosep, leftmargin=*]
  \item NeISF++, the first polarized inverse rendering pipeline with pBRDF supporting conductors and dielectrics.
  \item A novel geometry initialization approach using DoLP images, which is robust to strong specular reflections. 
  \item A real and synthetic multi-view polarimetric dataset consists of objects containing conductors and dielectrics.
\end{itemize}
The code and dataset will be made public upon acceptance.

\section{Related Works}
\subsection{Scene representations}
Exploring appropriate 3D representations is a crucial task for inverse rendering. Conventional renderers usually represent geometry as meshes and material as BRDFs and render the image via path tracing. However, these representations do not always work well for inverse rendering. For example, meshes are discrete data structures; the differentiability must be considered during the optimization. BRDF parameters are usually given by texture maps, which have resolution and sampling issues. Although path tracing creates photorealistic images, the computational cost is large. On the other hand, recent neural rendering works \cite{mildenhall2020nerf} show the possibility of representing 3D scenes using neural networks, which are compact and efficient. We mainly cover the three important aspects of inverse rendering, which are geometry, material, and lighting, and describe how to represent them using neural networks.

\noindent \textbf{Geometry}
The original NeRF \cite{mildenhall2020nerf} represents geometry as a density field. Although it shows stunning results in novel-view synthesis, the reconstructed geometry is noisy. On the other hand, some works like IDR \cite{yariv2020multiview} use SDF to represent the geometry and show a smooth surface estimation. Later works like VolSDF \cite{yariv2021volume}, NeuS \cite{wang2021neus}, or Objects as volumes \cite{miller2024objects} take advantage of both worlds by building connections between SDF and density via Laplace, logistic, or Gaussian functions. Another research branch is representing geometries as 3D \cite{kerbl20233d} or 2D \cite{huang20242d} Gaussians. Although these methods greatly accelerate the training and inference time, the reconstruction quality of geometry is still controversial \cite{huang20242d} compared to SDF-based methods. Therefore, we use VolSDF \cite{yariv2021volume} as our geometry representation. 

\noindent \textbf{Material}
Parametric BRDFs, which represent BRDF as a parametric model, are a commonly used method in conventional renderers. The parameters are usually provided via 2D texture maps, and UV mapping functions are used to project 2D points to 3D spaces. However, texture maps are discrete data, which may cause aliasing problems and thus require advanced sampling techniques. Recently, several neural rendering works \cite{boss2021nerd} have shown that the texture can be represented as a continuous 3D function using MLPs. Given the 3D position of the object's surface, the MLP directly returns the BRDF parameters without sampling. As for the selection of BRDF models, most existing neural rendering works \cite{zhang2022modeling, munkberg2022extracting, yao2022neilf, zhang2023neilf++, hasselgren2022shape, jin2023tensoir, wu2023nefii, cheng2023wildlight} use Disney BRDF \cite{burley2012physically}. However, this BRDF only describes the material property regarding light intensity. To describe the material property of polarization, pBRDFs \cite{baek2018simultaneous, hwang2022sparse, kondo2020accurate, ichikawa2023fresnel} are desired. Among them, Baek pBRDF \cite{baek2018simultaneous} is commonly used in polarized inverse rendering works. However, one limitation is that it only supports dielectric objects, which is problematic because conductors are also our target objects. We extend this pBRDF to also support conducts.

\noindent \textbf{Lighting}
Another important part of inverse rendering is the lighting representation. Early works only model the single-bounced light as a spherical Gaussian \cite{boss2021nerd, zhang2021physg}, low-resolution environment map \cite{zhang2021nerfactor, hasselgren2022shape}, or split-sum approximation \cite{boss2021neuralpil, munkberg2022extracting}. However, a strong limitation of only modeling the single-bounced is that their models can not handle phenomenons such as inter-reflection. 
Multi-bounce simulation is expensive and prone to noise when using the traditional Monte Carlo-based path tracing. Various solutions have been proposed, such as separating the direct and indirect light using a visibility mask \cite{zhang2022modeling, jin2023tensoir} or a bounding sphere \cite{liu2023nero}, and modeling them individually, pre-computing the multi-bounced light \cite{fipt2023}, only computing the last bounce explicitly, and using MLPs to record the rest bounces implicitly \cite{yao2022neilf, zhang2023neilf++, hadadan2021neural}. NeISF \cite{li2024neisf} is the first work that considers the multi-bounced polarized light by proposing the incident Stokes field. We follow this light representation.

\subsection{Polarization in inverse rendering}
Many works use polarization cues to disambiguate inverse rendering since polarization cues are sensitive to geometry and material. Single-view methods are mainly based on deep neural networks, and they have shown some progress in the inverse rendering of opaque \cite{deschaintre2021deep, duan2023end}, transparent \cite{shao2023transparent}, or translucent \cite{li2024deep} objects. However, the disadvantage of using data-driven methods is the lack of data for real-world objects. Using synthetic data to train the network may solve the problem of insufficient data, but it also introduces additional problems like domain gaps. Multi-view methods \cite{zhao2022polarimetric, azinovic2023high} are mainly based on analysis-by-synthesis, and the performance bottleneck comes from the aforementioned scene representation problem. 
Many works have recently attempted to combine neural representation and polarization cues. Early works mainly focus on the novel-view synthesis of polarimetric fields \cite{peters2023pcon} or Spectro-polarimetric fields \cite{kim2023neural}. Lately, PANDORA \cite{dave2022pandora} has been proposed for geometry reconstruction, and follow-up works have improved it by introducing tangent space consistency (MVAS \cite{cao2023multi}, NeRSP \cite{han2024nersp}) or polarimetric losses (PISR \cite{chen2024pisr}). NeISF \cite{li2024neisf} is the first neural rendering work that supports pBRDF decomposition, and it is the most similar work to ours. However, the strong limitation is that it only supports dielectrics. Our work can be considered a general version supporting conductors and dielectrics.

\subsection{Inverse rendering of specular objects}
The original NeRF \cite{mildenhall2020nerf} tends to create fake specular reflections by foggy effects. To solve the problem, ideas such as modeling a 3D varying refractive index \cite{bemana2022eikonal} and directional encoding \cite{verbin2022ref, ma2024specnerf} have been proposed. Later, some works attempt to reconstruct the accurate geometry of glossy objects by computing a reflection score map via an abnormal detector \cite{ge2023ref} or the difference between the albedo and intensity \cite{wang2024inverse}. Recently, several inverse rendering works also put their focus on reflective objects by considering the multi-bounce reflection \cite{liu2023nero},  integrating material priors \cite{lai2024glossygs}, and using a novel parallax-aware non-distant lighting representation \cite{cai2024pbir}. We show that the DoLP image is a powerful cue for the geometry initialization of glossy objects because the DoLP is invariant to light intensities and thus robust to the strong specular reflection.

\section{Preliminary}
We briefly introduce the mathematical representation of polarized light and material and recap NeISF \cite{li2024neisf}.

\subsection{Background of Polarization}
\noindent \textbf{Stokes vectors and Mueller matrices} are the mathematical descriptions of the polarized light and optical elements. When only linear polarization is considered, Stokes vectors $\mathbf{s} \in \mathbb{R} ^ 3$ have three elements $[s_0, s_1, s_2]$, where $s_0$ is the unpolarized light intensity, $s_1$ is the $0^{\circ}$ over $90^{\circ}$ linear polarization, and $s_2$ is the $45^{\circ}$ over $135^{\circ}$ linear polarization. Mueller matrices $\mathbf{M} \in \mathbb{R} ^ {3 \times 3}$ describe the polarimetric property of the material and are used to multiply Stokes vectors to represent the polarimetric light-object interaction.

\noindent \textbf{Baek pBRDF} \cite{baek2018simultaneous} is commonly used in the recent polarized inverse rendering works \cite{li2024neisf, dave2022pandora}. It separates the material model into diffuse and specular polarization. Diffuse polarization $\mathbf{M}_\text{dif}$ describes the process of light passing through the surface, undergoing multiple scattering events inside the object, leaving the object, and being captured by the sensor:
\begin{equation}
\label{equ:M_dif}
    \mathbf{M}_\text{dif} = (\frac{\text{a}}{\pi}\cos{\theta_i}) \mathbf{F}^T_o \cdot \mathbf{D} \cdot \mathbf{F}^T_i,
\end{equation}
where $\text{a}$ is the diffuse albedo, $\theta_{i,o}$ denotes the incident / outgoing angle, $\mathbf{D} \in \mathbb{R}^{3 \times 3}$ is a depolarizer, and $\mathbf{F}^T_{i,o} \in \mathbb{R}^{3 \times 3}$ is the Fresnel transmission term. Specular polarization $\mathbf{M}_\text{spec}$ comes from the single-bounced mirror reflection:
\begin{equation}
\label{equ:M_spec}
    \mathbf{M}_\text{spec} = k_s \frac{DG}{4\cos{\theta_o}} \mathbf{F}^R,
\end{equation}
where $k_s$ is the specular coefficient, $D$ is the GGX distribution function \cite{walter2007microfacet},  $G$ is the Smith function, and $\mathbf{F}^R \in \mathbb{R}^{3 \times 3}$ is the Fresnel reflection term.

\subsection{Recap of NeISF}
\noindent \textbf{Polarimetric rendering equation}
NeISF \cite{li2024neisf} uses Baek pBRDF as the material model. As a result, the corresponding Rendering Equation \cite{kajiya1986rendering} is:
\begin{equation}
    \label{equ:rendering equation neisf}
    \mathbf{s} =  \int_\Omega \mathbf{R}^\text{cam}_\text{dif}  \cdot \mathbf{M}_\text{dif}   \cdot \mathbf{s}^\text{r}_\text{dif}  +  \mathbf{R}^\text{cam}_\text{spec}  \cdot \mathbf{M}_\text{spec}    \cdot \mathbf{s}^\text{r}_\text{spec} \,d \bm{\omega}_i,
\end{equation}
where $\mathbf{s}$ is the final Stokes vector captured by the sensor, $\mathbf{R}^\text{cam}$ is the rotation Mueller matrix which rotates the outgoing Stokes vector to the camera's reference frame, $\mathbf{s}^\text{r}$ is the incident Stokes vector which is already rotated to the reference frame of the material Mueller matrix $\mathbf{M}$, $\bm{\omega}_i$ is the incident direction, and the subscript $_\text{dif/spec}$ denotes the diffuse/specular term. The integral of Eq. \ref{equ:rendering equation neisf} is solved by a Fibonacci sphere sampling.

\noindent \textbf{Neural fields}
They represent the geometry as an SDF $\mathbb{S}:\mathbf{x}\rightarrow d$, where $\mathbf{x}$ is the 3D position, $d$ is the distance, and the normal $\mathbf{n}$ is obtained by computing the gradient of the SDF. The material is represented as a BRDF field $\mathbb{B}:\mathbf{x} \rightarrow \{r, \textbf{a}\}$, where $r$ is the roughness and $\textbf{a}$ is the diffuse albedo. The lighting is represented as an incident Stokes field $\mathbb{L}:\{\mathbf{x}, \bm{\omega}_i\}\rightarrow \{ \mathbf{s}^\text{r}_\text{dif}, \mathbf{s}^\text{r}_\text{spec} \}$. Then Eq. \ref{equ:rendering equation neisf} is used for rendering the Stokes vector $\mathbf{s}$, and the entire model training is self-supervised by the GT Stokes vectors.

\section{NeISF++}
Given multi-view polarized images of objects containing dielectrics and conductors, we reconstruct geometry and material simultaneously. The reconstructed 3D assets are compatible with conventional renderers that are capable of downstream tasks such as relighting and material editing. The sections are arranged as follows: We introduce the material model that is designed for both conductors and dielectrics in Sec. \ref{sec:material model}. The training is divided into two stages: the first stage is the geometry initialization using DoLP images (Sec. \ref{sec:geometry init}), and the second stage is the joint optimization  (Sec. \ref{sec:joint opt}) of geometry, material, and lighting using the proposed material model.

\begin{figure}[t]
  \centering
  \includegraphics[width=\linewidth]{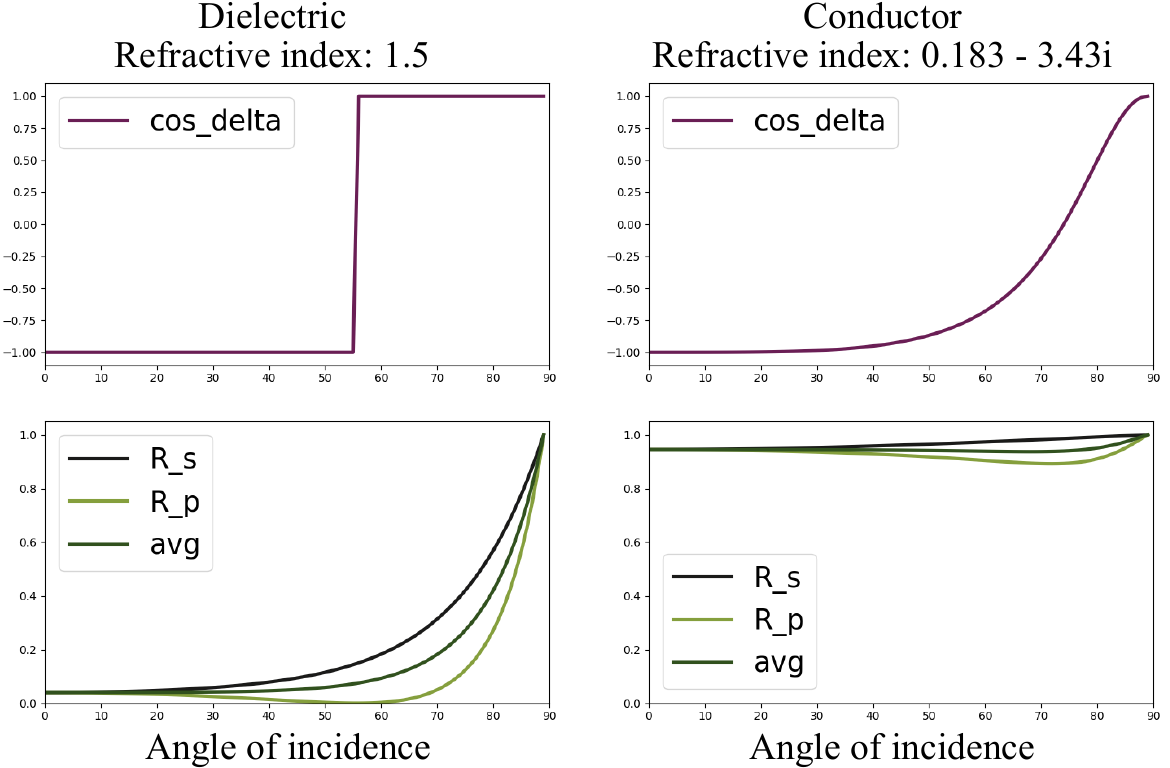}
  \caption{Cosine values of phase delay (upper) and reflection coefficients (bottom) of a typical conductor (gold at 633nm) and dielectric (refractive index equals 1.5). ``R\_s" and ``R\_p" are the perpendicular and parallel components of reflectance, ``avg" denotes their average value. }
  \label{fig:conductor_dielectric_ior}
\end{figure}

\subsection{Material model}
\label{sec:material model}
As briefly mentioned before, the previous polarized inverse rendering works \cite{li2024neisf, dave2022pandora} use Baek pBRDF as their material model, which only supports dielectrics. Extending this pBRDF also to support conductors requires solving some discrepancies. Because conductors are practically opaque, \cite{collett2005field}, subsurface scattering, which is the key factor creating diffuse polarization, does not exist for conductors. We propose a simple yet efficient solution by adding a binary indicator $m$ before the diffuse polarization term. As a results, Eq. \ref{equ:rendering equation neisf} should be rewritten as:
\begin{equation}
    \label{equ:rendering equation neisfpp}
    \mathbf{s} =  \int_\Omega m \mathbf{R}^\text{cam}_\text{dif}  \cdot \mathbf{M}_\text{dif}   \cdot \mathbf{s}^\text{r}_\text{dif}  +  \mathbf{R}^\text{cam}_\text{spec}  \cdot \mathbf{M}_\text{spec}    \cdot \mathbf{s}^\text{r}_\text{spec} \,d \bm{\omega}_i.
\end{equation}
The value of $m$ is either set to 0 for conductors or 1 for dielectrics. Note that, unlike the metallic term used in Disney BRDF \cite{burley2012physically} is a continuous parameter, for physical realism, $m$ should be a discrete parameter. Even if the existence of the diffuse polarization problem has been solved by the binary indicator, the specular polarization term is still different for conductors and dielectrics. Concretely, the refractive index for dielectrics is real numbers, and the refractive index for conductors is complex numbers. This causes the reflection coefficients and the phase delay of the Fresnel reflection term $\textbf{F}^R$ to be dramatically different for conductors and dielectrics (See Fig. \ref{fig:conductor_dielectric_ior} for reference). Unfortunately, $\textbf{F}^R$ in Baek pBRDF does not support complex numbers:
\begin{equation}
    \mathbf{F}^R = \begin{bmatrix}
R^+ & R^- & 0  \\
R^- & R^+ & 0  \\
0 & 0 & R^{\times} \cos{\Delta}
\end{bmatrix},
\end{equation}
because the calculation of reflection coefficients $R^+, R^-$ is designed for real numbers, and the phase delay $\Delta$ is hard coded to $\pi$ or $0$ when the incident angle is less or larger than the Brewster angle. We calculate the reflection coefficients and phase delay based on the Fresnel wave theory \cite{collett2005field}, so the  Fresnel reflection term supports a complex refractive index (Details can be found in the supplementary document).

\noindent \textbf{Assumptions} Theoretically, the binary indicator $m$ can be optimized jointly just like the other pBRDF parameters. However, adding parameters to be optimized significantly increases the ambiguity problem. Therefore, we assume $m$ is given by the user-specified conductor-dielectric mask during the training to reduce the difficulty of the task. In addition, we only estimate the refractive index for the conductor part, and for the dielectric part, we follow NeISF \cite{li2024neisf} to assume it as 1.5.

\begin{figure}[t]
  \centering
  \includegraphics[width=0.8\linewidth]{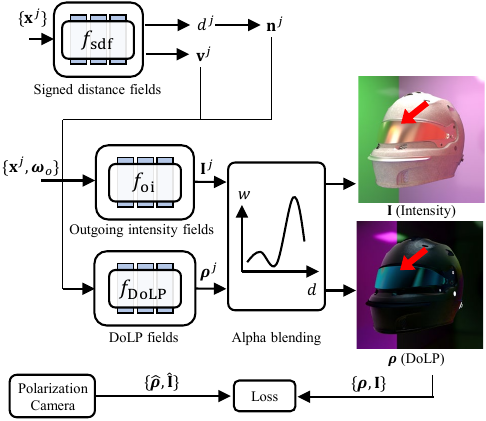}
  \caption{Geometry initialization pipeline using both intensity and DoLP images. Intensity on the conductor area suffers from strong specular reflections, while DoLP images are less affected by strong specular reflections. The initialized signed distance fields $f_\text{sdf}$ will continue to be trained in the joint optimization stage.}
  \label{fig:overview_init}
\end{figure}

\begin{figure}[t]
  \centering
  \includegraphics[width=\linewidth]{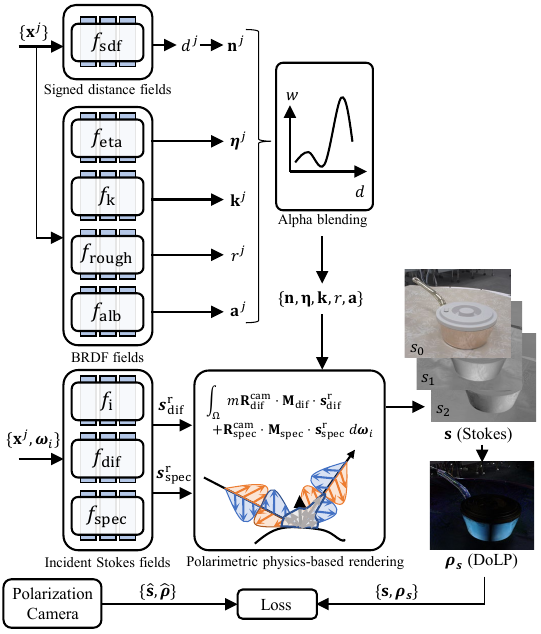}
  \caption{Overview of the joint optimization stage.}
  \label{fig:overview_joint}
\end{figure}

\subsection{Geometry initialization}
\label{sec:geometry init}

In this section, we demonstrate that DoLP is a suitable image space for geometry initialization, especially when strong specular reflections exist. To stabilize the training, previous inverse rendering works \cite{zhang2023neilf++} usually start the training with geometry initialized by volume rendering techniques \cite{yariv2021volume}. However, a well-known problem with these methods is their bad performance for reflective objects. The challenging part of reflective objects is their view-dependent appearance property. Although these techniques have already modeled the view-dependent appearance property by inputting the view direction to MLPs, the inherent ambiguity of inverse rendering usually leads to wrong geometry reconstruction. This wrong geometry reconstruction behavior will be escalated by specular reflections with strong intensities. For example, the reflected light directly comes from the light source, which is commonly observed daily. We propose using DoLP to train a VolSDF \cite{yariv2021volume} to initialize the geometry:
\begin{equation}
    f_\text{DoLP}(\mathbf{x}^j, \bm{\omega}_o, \mathbf{n}^j, \mathbf{v}^j  ) = \bm{\rho}^j, 
\end{equation}
where $f_\text{DoLP}$ is an MLP, $\mathbf{x}^j, \mathbf{n}^j, \mathbf{v}^j,  \bm{\rho}^j$ are the position, normal, feature vector, and DoLP at the sampled point along the ray. $\bm{\omega}_o$ is the outgoing direction. The normal and feature vectors are given by the SDF net $f_\text{sdf}$. The DoLP of the object surface is computed via alpha-blending $\bm{\rho} = \sum_{j=1}^N w^j \bm{\rho}^j$, and $w^j$ is the weight calculated by the volume rendering. The loss is computed by the $L_1$ loss between the GT DoLP and the estimated one. The aforementioned strong specular reflection intensity problem does not exist in DoLP images because it is invariant to light intensity (See Fig. \ref{fig:overview_init} for reference). Besides, DoLP is strongly related to geometry, making it suitable for geometry initialization. To stabilize the training, we still use the intensity images to compute the loss, just like VolSDF \cite{yariv2021volume}. However, we only assign the intensity loss a tiny weight. As a result, the loss function is:
\begin{equation}
    L_\text{init} = \lambda_{\bm{\rho}}L_1(\bm{\rho}, \hat{\bm{\rho}}) + \lambda_{\textbf{I}} L_1(\textbf{I}, \hat{\textbf{I}}) + \lambda_\text{Eik} L_\text{Eik}, 
\end{equation}
where $\hat{\textbf{I}}, \hat{\bm{\rho}} $ are the GT intensity and DoLP images, $\textbf{I}, \bm{\rho}$ are the reconstructed intensity and DoLP images. $L_\text{Eik}$ is the Eikonal regularization \cite{gropp2020implicit} to avoid the everywhere zero solution of the SDF net. 

\subsection{Joint optimization}
\label{sec:joint opt}
After the geometry initialization in Sec. \ref{sec:geometry init}, we jointly optimize the geometry, material, and lighting (Fig. \ref{fig:overview_joint}). Because the overall structure follows NeISF \cite{li2024neisf}, we only focus on the key contribution of this work, and the entire model description can be found in the supplementary material. We represent geometry as an SDF, material as a BRDF field, and lighting as an incident Stokes field.  The main difference comes from the BRDF field. Unlike common dielectrics such as acrylic glass (1.49), polypropylene plastic (1.49), and quartz (1.458), have similar refractive index $\bm{ior}$, the $\bm{ior}$ of conductors varying significantly among common conductors such as aluminum, copper, steel. In addition, the $\bm{ior}$ of conductors are spectrally varying, affecting the object's appearance a lot. Therefore, it is impossible to assume the $\bm{ior}$ as a constant for conductors. To solve this problem, we estimate the $\bm{ior}$ in the BRDF field as well. The $\bm{ior}$ of conductors is a complex number, and it can be represented as follows: $\bm{ior} = \bm{\eta} - \bm{k}i$, where $\bm{\eta}\in \mathbb{R} ^ 3$ is the real part and $\bm{k}\in \mathbb{R} ^ 3$ is the imaginary part. Therefore the BRDF field should be rewritten as: $\mathbb{B}:\mathbf{x} \rightarrow \{r, \textbf{a}, \bm{\eta}, \bm{k}\}$. After estimating $\bm{\eta}$ and $\bm{k}$, we combine and convert them to complex numbers.  As for the dielectric part, NeISF \cite{li2024neisf} shows an impressive result even with $\bm{ior}$ assumed as a constant. Following their work, we also assume the $\bm{ior}$ of dielectrics as $1.5$. After estimating all parameters, the final Stokes vectors are rendered by Eq. \ref{equ:rendering equation neisfpp}. Note that the rendering equation should be repeated three times for rendering RGB channels. We compute $L_1$ loss between the reconstructed Stokes vectors and the GT. In addition, we also reconstruct the DoLP image: $\bm{\rho}_s = \sqrt{\textbf{s}[1]^2 + \textbf{s}[2]^2} / {\textbf{s}[0]}$, and the loss function is:
\begin{equation}
    L_\text{joint} = \lambda_{\bm{\rho}_s}L_1(\bm{\rho}_s, \hat{\bm{\rho}}) + \lambda_{\textbf{s}} L_1(\textbf{s}, \hat{\textbf{s}}) + \lambda_\text{Eik} L_\text{Eik},
\end{equation}
where $\textbf{s}, \hat{\textbf{s}}$ are the reconstructed and GT Stokes vectors. We utilize the auto-differentiation of complex numbers feature in PyTorch \cite{paszke2017automatic} so that the entire pipeline is differentiable.

\section{Experiments}
\subsection{Datasets}
Although many polarized multi-view datasets exist, none contain both conductors and dielectrics. In addition, most of the existing datasets are LDR, which may suffer from unknown gamma correction and saturation problems. We propose real and synthetic multi-view polarized HDR datasets containing conductors and dielectrics. For the synthetic dataset, we rendered images using Mitsuba 3 \cite{Jakob2020DrJit}. We rendered 110 views for each scene with GT Stokes vectors, DoLP images, object masks, conductor-dielectric masks, diffuse albedo maps, roughness maps, and refractive index maps. We split the dataset and used 100 views for training and 10 for evaluation. For the real dataset, we captured images with a polarization camera (FLIR BFS-U3-51S5PC-C). Because the camera only supports LDR capture, we captured images with different exposure times and composited them to obtain HDR images. For each scene, we captured roughly 100 views for training and 2-4 for evaluation. For each view, we captured four linearly polarized images, which are used to calculate the Stokes vectors and DoLP images. We also manually created object masks and conductor-dielectric masks using Photoshop \cite{adobephotoshop}.

\subsection{Baselines}
To our knowledge, this is the first work focusing on the polarized inverse rendering of both conductors and dielectrics, so looking for competitors with exactly the same target is difficult. As a result, \textbf{NeISF} \cite{li2024neisf} is the second best choice. Although it does not support the refractive index estimation of conductors, the geometry, roughness, diffuse albedo, and relighting results can be compared. In addition, although some works \cite{dave2022pandora, cao2023multi, chen2024pisr} do not support material estimation, they still use polarization for geometry reconstruction. We selected the most representative work \textbf{PANDORA} \cite{dave2022pandora} as our competitor. On the other hand, although some works do not use polarization cues, they are specifically designed for specular object reconstruction. Among them, we chose the most influential work \textbf{NeRO} \cite{liu2023nero} to compare the geometry reconstruction. Finally, we also used a famous volume rendering work \textbf{VolSDF} \cite{yariv2021volume} as the competitor. We also conducted an ablation study. Instead of using the geometry initialization method proposed in Sec. \ref{sec:geometry init} (we name it \textbf{VolSDF-DoLP}), we used the original VolSDF \cite{yariv2021volume} to initialize the geometry and train the joint optimization stage. We call this ablated version \textbf{Ours-a}.

\begin{figure*}[t]
  \centering
  \includegraphics[width=\linewidth]{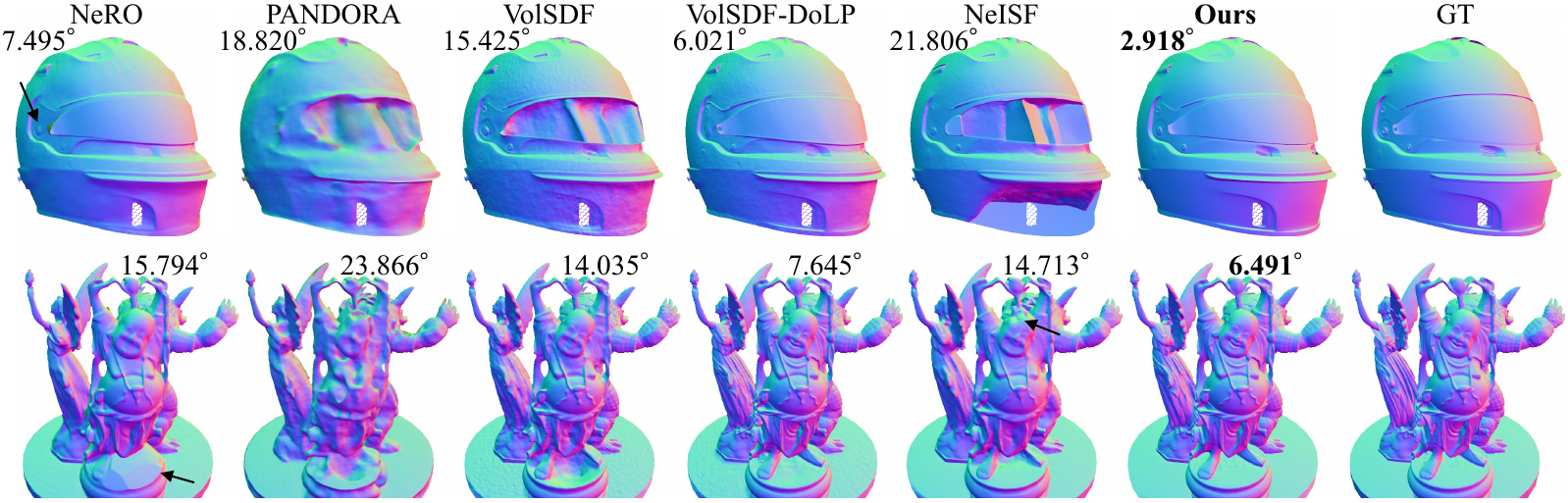}
  \caption{Surface normal results of synthetic data. Mean angular errors are on the top. Our method shows a better reconstruction quality than NeRO \cite{liu2023nero} and PANDORA \cite{dave2022pandora}. NeISF \cite{li2024neisf} failed because of the wrong material model and the poor geometry initialization of VolSDF \cite{yariv2021volume}. Our geometry initialization method VolSDF-DoLP shows a better reconstruction quality.   }
  \label{fig:normal_synthetic}
\end{figure*}

\begin{figure*}[t]
  \centering
  \includegraphics[width=\linewidth]{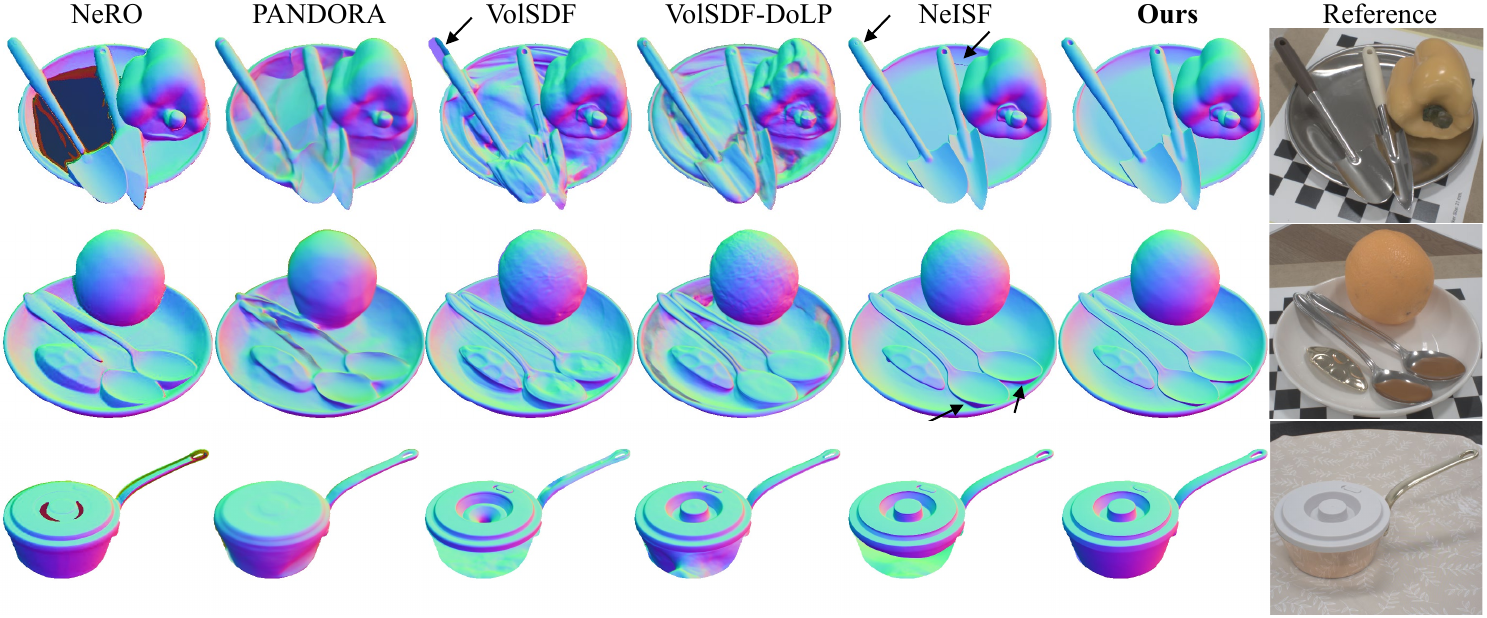}
  \caption{Surface normal reconstruction results of real data. }
  \label{fig:normal_real}
\end{figure*}

\begin{table*}[ht]
\begin{center}
\begin{tabular}{*8l}
\hline
{}  & Ours & Ours-a & VolSDF-DoLP & VolSDF \cite{yariv2021volume} & NeISF \cite{li2024neisf} & PANDORA\cite{dave2022pandora} & NeRO\cite{liu2023nero} \\
\hline
Stanford scan & 6.487$^{\circ}$ & \textbf{6.480}$^{\circ}$ & 7.641$^{\circ}$ & 11.754$^{\circ}$ & 14.022$^{\circ}$ & 21.740$^{\circ}$ & 13.352$^{\circ}$ \\
                
Helmet & \textbf{1.789}$^{\circ}$ & 2.400$^{\circ}$ & 4.715$^{\circ}$ & 8.829$^{\circ}$ & 10.303$^{\circ}$ & 13.212$^{\circ}$ & 5.001$^{\circ}$\\
                
\hline
\end{tabular}
\end{center}
\caption{Surface normal reconstruction results on our synthetic dataset. Mean angular errors are computed on the average of 10 test views.}
\label{table:compare_normal}
\end{table*}

\begin{table}[ht]
\begin{center}
\begin{tabular}{*5l}
\hline
{}  & {}& Ours & Ours-a & NeISF \cite{li2024neisf}\\
\hline
\multirow{4}{*}{Stanford scan} 
                
                & Roughness &  \textbf{.0706} & .0821 & .2425\\
                & Albedo & \textbf{.0468} & .1289 & .2954\\
                & Eta & \textbf{.0685} & .0722 & -\\
                & K & \textbf{.4300} & .5107 & -\\
                
\hline
\multirow{4}{*}{Helmet} 
                
                & Roughness  & \textbf{.0161} & .0199 & .2075\\
                & Albedo  & \textbf{.0615} & .0716 & .4467\\
                & Eta & \textbf{.0717} & .1937 & -\\
                & K & \textbf{.6526} & 1.327 & -\\
\hline
\end{tabular}
\end{center}
\caption{Material reconstruction results on the proposed synthetic dataset. We compute the mean absolute errors on 10 test views for all material parameters. We count the errors of Eta and K for the conductor part and albedo for the dielectric part. Roughness errors are computed over the entire object.}
\label{table:compare_material}
\end{table}

\subsection{Results}

\begin{figure*}[t]
  \centering
  \includegraphics[width=\linewidth]{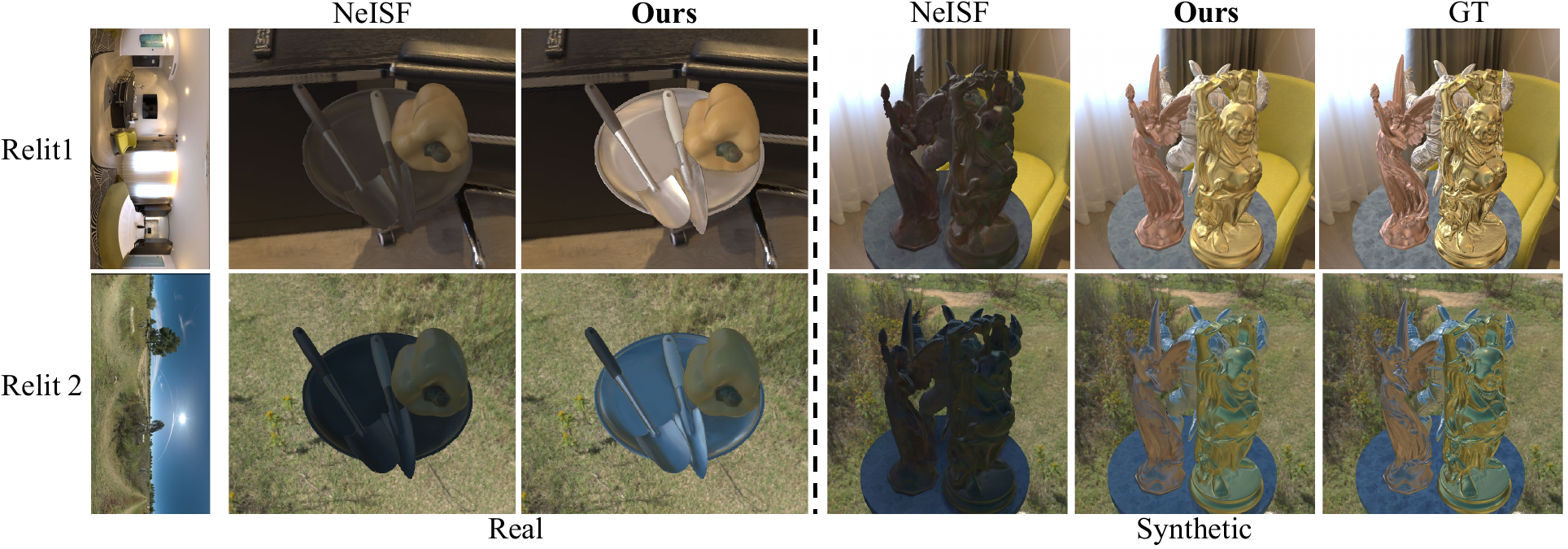}
  \caption{Relighting results. Even though NeISF \cite{li2024neisf} can reconstruct plausible geometry, the relighting result for the conductor part is not realistic due to the inaccurate material model. While our results are shiny and similar to the GT.}
  \label{fig:relit}
\end{figure*}

\begin{figure}[t]
  \centering
  \includegraphics[width=\linewidth]{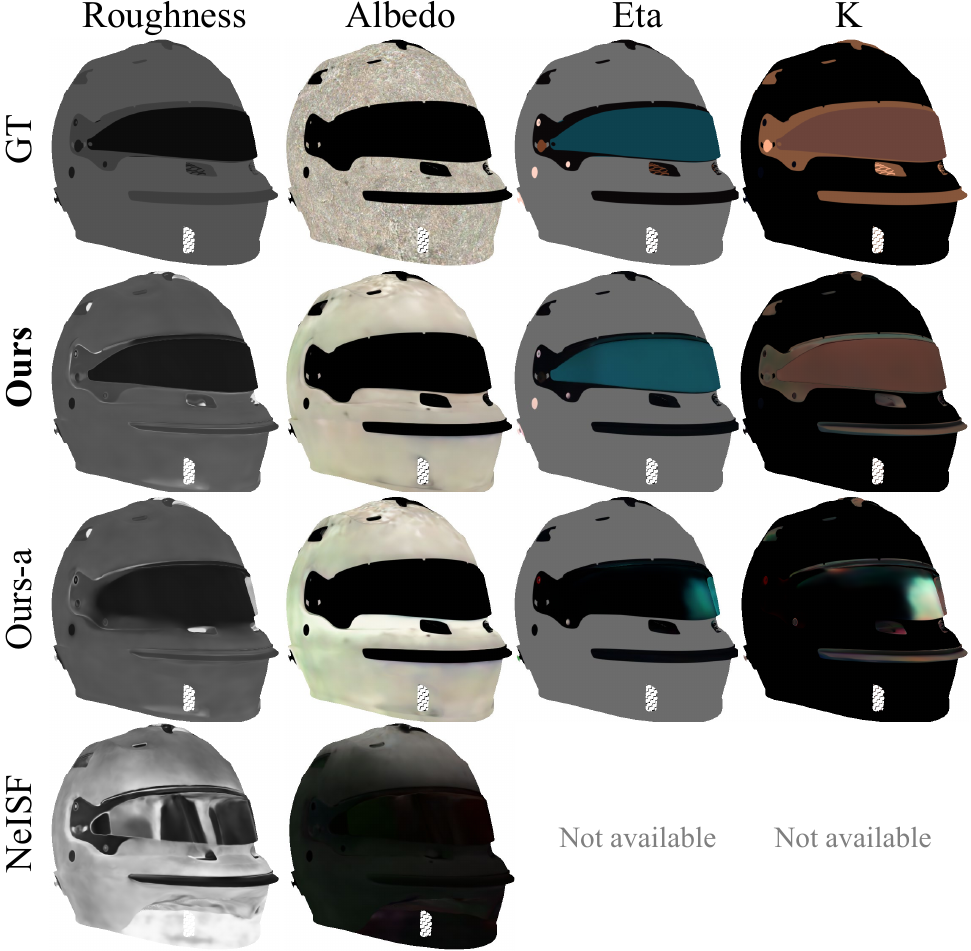}
  \caption{Material reconstruction result. Eta and K are the real and imaginary components of the complex refractive index. We normalize them to 0-1 for visualization. NeISF \cite{li2024neisf} completely failed to reconstruct the material due to the wrong geometry reconstruction. ``Ours" are significantly better than ``Ours-a", especially for the refractive index reconstruction. }
  \label{fig:material_syn}
\end{figure}

\noindent \textbf{Geometry reconstruction} 
We compare surface normal results for synthetic data qualitatively (Fig. \ref{fig:normal_synthetic}) and quantitatively (Tab. \ref{table:compare_normal}). Due to the lack of GT data for the real data, we only provide qualitative results in Fig. \ref{fig:normal_real}.

\noindent \textbf{Material reconstruction}
We report the qualitative results of material reconstruction in Fig. \ref{fig:material_syn} and quantitative results in Tab. \ref{table:compare_material}. Although NeRO \cite{liu2023nero} also supports material estimation, the BRDF model differs from ours. Therefore, it is meaningless to compare the reconstructed BRDF parameters. NeISF \cite{li2024neisf} does not support the complex refractive index reconstruction, thus we only compare the albedo and roughness. In addition, we also compare with ``Ours-a".

\noindent \textbf{Relighting} We demonstrate one important downstream task of inverse rendering: relighting. We compare the results between our method and NeISF \cite{li2024neisf} in Fig. \ref{fig:relit}. 

\section{Limitations and Future Works}
\label{sec:limitation}
\noindent \textbf{Conductor-dielectric masks} are assumed as known parameters in this work. However, manually creating the masks is time-consuming and labor-intensive, which may limit the method's practicality. Therefore, the automatic generation of this mask is desired. We discuss two potential solutions to this problem, which are data-driven and error-driven. For the data-driven approach, one can consider mask generation as a material segmentation task \cite{liang2022multimodal, sifnaios2024deep} through training a neural network on a large labeled dataset. For the error-driven approach, one can optimize a material mask to separate conductors and dielectrics according to the rendering errors, which have been verified effective for separating emitters and non-emitters \cite{fipt2023}. 

\section{Conclusion}
We have proposed NeISF++, a polarized inverse rendering pipeline that supports both conductors and dielectrics. It relies on the following novelties. The first one is a general pBRDF, which describes both conductors and dielectrics. The second one is a novel geometry initialization method using DoLP images. With these two novelties, NeISF++ outperforms the existing inverse rendering models for geometry and material decomposition on both the proposed synthetic and real-world datasets. The relighting comparison between our method and NeISF \cite{li2024neisf} also shows the importance of correctly modeling conductors. However, several limitations mentioned in Sec. \ref{sec:limitation} still exist and are worth further exploration.

{\small
\bibliographystyle{ieee_fullname}
\bibliography{egbib}
}

\end{document}